\documentclass{article}
\usepackage{spconf,amsmath,epsfig}
\usepackage{booktabs}
\usepackage{multirow}
\usepackage{siunitx}
\usepackage{hyperref}
\usepackage{txfonts}

\title{Transformer-based SAR image Despeckling}

\name{Malsha V. Perera, Wele Gedara Chaminda Bandara, Jeya Maria Jose Valanarasu, and Vishal M. Patel \thanks{\copyright \; 2022 IEEE. Personal use of this material is permitted. Permission from IEEE must be obtained for all other uses, in any current or future media, including reprinting/republishing this material for advertising or promotional purposes, creating new collective works, for resale or redistribution to servers or lists, or reuse of any copyrighted component of this work in other works.}}
\address{Johns Hopkins University\\
Department of Electrical and Computer Engineering\\
\{jperera4, wbandar1, jvalana1, vpatel36\}@jhu.edu}
\begin{document}
%
\maketitle
\begin{abstract}
Synthetic Aperture Radar (SAR) images are usually degraded by a multiplicative noise known as speckle which makes processing and interpretation of SAR images difficult. In this paper, we introduce a transformer-based network for SAR image  despeckling. The proposed despeckling network comprises of a transformer-based encoder which allows the network to learn global dependencies between different image regions - aiding in better despeckling. The network is trained end-to-end with synthetically generated speckled images using a composite loss function. Experiments show that the proposed method achieves significant improvements over traditional and convolutional neural network-based despeckling methods on both synthetic and real SAR images. Our code is available at : \url{https://github.com/malshaV/sar_transformer}
\end{abstract}
\setlength{\belowdisplayskip}{0pt} \setlength{\belowdisplayshortskip}{0pt}
\setlength{\abovedisplayskip}{0pt} \setlength{\abovedisplayshortskip}{0pt}

\section{Introduction}
\label{sec:intro}
Synthetic Aperture Radar (SAR), like other coherent imaging systems,  is also affected by speckle which is a signal-dependent, spatially correlated noise. The presence of speckle impairs downstream tasks of SAR images such as segmentation, recognition, and detection. Therefore, despeckling of SAR images positively impacts these downstream tasks.

In the past few decades, different despeckling approaches have been proposed in the literature. Filter-based approaches used for despeckling can be generally categorized as local and non-local filters. Lee filter \cite{lee_filter} and Kuan filter \cite{kuan_filter} are some despeckling methods which use local filters, while PPB \cite{ppb} and SAR-BM3D \cite{sar_bm3d} are examples for non-local filter-based despeckling methods.  A survey of different SAR image despecking methods can be found in \cite{Despecking_Survey_2021}.

With the advent of deep learning, significant advances have been made in SAR despeckling. Unlike the despeckling approaches mentioned earlier, convolutional neural networks (CNNs) require a pair of noisy and clean (ground truth) images so that they can be trained in a supervised way. However, clean references for SAR images do not exist. Therefore, supervised CNN-based methods generate reference images via synthetic speckle generation or multi-temporal fusion. SAR-CNN \cite{SARCNN} is a CNN-based despeckling network which is trained on SAR images transformed to the hormomophic form. The ground truth for SAR-CNN is generated via multi-temporal fusion. Wang \textit{et al.} \cite{IDCNN} proposed ID-CNN which directly estimates the noise in the original domain and is trained on synthetic SAR images. Here, the despeckled image is obtained by dividing the speckled SAR image by the estimated noise. Despeckling approaches such as \cite{KLD,MRDDANet}  proposed slight variations on the above CNN-based methods by introducing different architectures and loss functions. 

With the success of transformers in natural language processing, they have been successfully adopted  to many vision tasks in recent years. Dosovitskiy \textit{et al.} \cite{vit} proposed Vision Transformer (ViT) which resulted in an impressive image classification performance on ImageNet as transformers have the ability to learn long-range dependencies in an image. In ViT, the input image is split into multiple linearly embedded patches which are fed into a transformer encoder. Following ViT, many studies have employed transformers in various computer vision tasks achieving state-of-the-art performance. For example, U-former \cite{uformer} and SwinIR \cite{swinir} are transformer-based methods proposed for image restoration. To the best of our knowledge, transformer-based SAR despeckling has not been studied in the literature.

To this end, we propose a transformer-based network for SAR image despeckling. The proposed network consists of a transformer encoder and a CNN decoder. Unlike ViT which only produces feature maps with fixed resolutions, the hierarchical nature of our proposed transformer encoder allows us to generate high resolution fine features as well as low resolution coarse features required for SAR image despeckling. Similar to \cite{segformer}, our transformer encoder avoids interpolating positional codes when performing inference. Therefore, the encoder can easily adapt to different test resolutions without impairing the performance. The proposed network is trained end-to-end with synthetically speckled optical images. Finally, we compare the performance of our proposed network on synthetic  images as well as real SAR images with several non-local filter and CNN-based methods where we achieve state-of-the-art performance.

\begin{figure*}[htbp]
\centerline{\includegraphics[width=.85\textwidth]{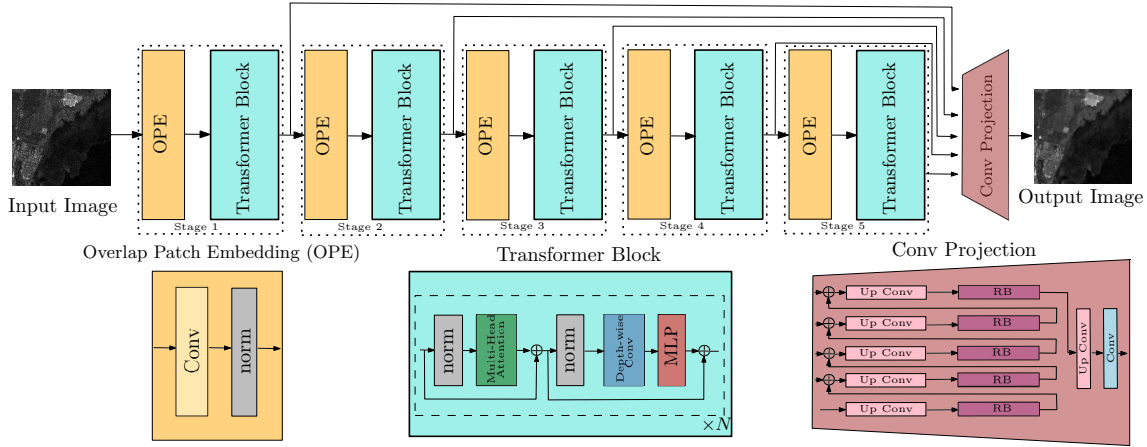}}
\caption{Overview of the proposed despeckling transformer.}
\label{network}
\end{figure*}

\section{Proposed Method}
\label{sec:method}

\subsection{Speckle in SAR}
\label{speckle simulation}
For a SAR image with an average number of $L$ looks, the observed SAR intensity $y$ is related to the speckle-free SAR intensity $x$, as follows:
\begin{equation}
   y = xn,
\label{eq1}
\end{equation}
where $n$ is the multiplicative speckle. Under the hypothesis of fully developed speckle, $n$ follows a Gamma distribution with unit mean and a variance of $1/L$. Therefore, the probability distribution of $n$ is given by,
\begin{equation}
    p(n) = \frac{1}{\Gamma(n)}L^Ln^{L-1}e^{-Ln},
\label{eq2}
\end{equation}
where $\Gamma(.)$ is the Gamma function. Given $y$, our goal is to estimate $x$.

\subsection{Network architecture}
The proposed network architecture is illustrated in Fig. \ref{network}. The noisy input image $y$  is passed through an encoder comprising of 5 stages, each stage containing an overlap patch embedding block and a transformer block. The last stage of the encoder is followed by a convolutional projection block that acts as a decoder. At each encoder stage, the resolution of the input is halved to allow the transformer to learn both coarse and fine details. Output from each encoder stage (except for the final stage) is then fed to both the next encoder stage and the convolutional projection block. The output of the final encoder stage is passed only to the convolutional projection block. The components of the network are described in detail below.\\

\noindent \textbf{Overlap Patch Embedding Block:}
Input to each encoder stage is first passed through an overlap patch embedding (OPE) block which uses the overlap patch merging process introduced in \cite{segformer}. Its purpose is to combine overlapping feature patches to obtain features of the same
size as that of non-overlapping patches before passing the features to the transformer block. In order to obtain the overlapped feature patches, the input to OPE block is passed through a convolutional layer of kernel size $k$, embedded dimensions (i.e. number of filters) $e$, stride $s$, and padding $p=k/2$. In this work, we set $s=2$ for all the OPE blocks in the network. The $e$ values were set to 32, 64, 128, 320, 512 and $k$ values were set equal to 7, 3, 3, 3, 3 in each OPE block of stages 1 to 5, respectively. Next, the flattened output of the convolutional layer is followed by layer normalization.

\noindent \textbf{Transformer Block.}
The output of each transformer block $\text{T}(\mathbf{I}_{in})$ with respect to an input $\mathbf{I}_{in}$ can be summarized as follows:
\begin{equation}
    \text{T}(\mathbf{I}_{in}) = \text{MLP}(\text{DWC}(\text{X}(\mathbf{I}_{in}))) + \text{X}(\mathbf{I}_{in})
\end{equation}
where, 
\begin{equation}
    \text{X}(\mathbf{I}_{in}) = \text{MHA}(\mathbf{I}_{in}) + \mathbf{I}_{in},
\end{equation}
where, $\text{MHA}$, $\text{DWC}$ and $\text{MLP}$ correspond to multi-head attention layer, depth-wise convolution and multi-layer perceptron, respectively. Note that we perform layer normalization on $\mathbf{I}_{in}$ and $\text{X}(\mathbf{I}_{in})$ before passing those to $\text{MHA}$ and $\text{DWC}$, respectively.
In the original multi-head self-attention process, the queries $\mathbf{Q}$, keys $\mathbf{K}$ and values $\mathbf{V}$ have the same dimensions $d$, and the self-attention is calculated as:
\begin{equation}
    \text{Attention}(\mathbf{Q}, \mathbf{K}, \mathbf{V}) = \text{Softmax} \Big( \frac{\mathbf{Q}\mathbf{K}^\top}{\sqrt{d}} \Big ) \mathbf{V}
\end{equation}
In the proposed network, each transformer block in stages 1 to 5 has 1, 1, 2, 4, and 8 attention heads, respectively. To reduce the computational complexity, we use the reduction ratio $R$ introduced in \cite{pyramidvt}, and we set $R = 2$ for all the transformer blocks in the network. The self-attention features are then passed through the depth-wise convolutional block that provides positional information for transformers \cite{segformer}. Subsequently, the output from the depth-wise convolution block is sent through a Gaussian error linear unit (GELU) before passing it through an MLP that comprises of a dropout layer and a linear layer. The output size of the MLPs in stages 1 to 5 are set to 32, 64, 128, 320, and 512, respectively.\\

\noindent \textbf{Convolutional Projection Block.}
The convolutional projection block is used to upsample the outputs from the transformer blocks to the original image size as illustrated in Fig. \ref{network}. The upsampling layers increase the resolution by a factor of two. The output of the residual block ($\text{RB}$) for a given input $\mathbf{I}_{in}$ can be computed as follows:
\begin{equation}
    \text{RB}(\mathbf{I}_{in}) = \text{Conv}_{3 \times 3}(\text{ReLU}( \text{Conv}_{3\times3}(\mathbf{I}_{in}))) + \mathbf{I}_{in},
\end{equation}
where $\text{Conv}_{3\times3}$ refers  to  $3 \times 3$ convolution layer and $\text{ReLU}$ denotes rectified linear unit.

\subsection{Loss Function}
We train the proposed network using the $l_2$ loss ($\mathcal{L}_{l_2}$) given by: 
\begin{equation}
    \mathcal{L}_{l_2} = \| \hat{x} - x \|_2^2,
\end{equation}
where $x$ and $\hat{x}$ denote the ground truth and the predicted output, respectively. In addition, we employ total variation loss in order to encourage smoothness while preserving edges. The total variation loss ($\mathcal{L}_{tv}$) is defined as follows
\begin{equation}
    \mathcal{L}_{tv} = \sum_{i,j} |\hat{x}_{i+1,j} - \hat{x}_{i,j}| + | \hat{x}_{i,j+1} - \hat{x}_{i,j}|.
\end{equation}
The overall loss function $\mathcal{L}$ is given by:
\begin{equation}
    \mathcal{L} = \lambda_1 \mathcal{L}_{l_2} + \lambda_2 \mathcal{L}_{tv},
\end{equation}
where $\lambda_1, \lambda_2$ are the weights that specify the contribution of $\mathcal{L}_{l_2}$ and $\mathcal{L}_{tv}$, respectively. In this study, we set $\lambda_1 = 1$ and $\lambda_2 = 5 \times 10^{-5}$ to put a strong emphasis on the $\mathcal{L}_{l_2}$ loss.

\section{Experiments and Results}
\label{sec:experiments}
To train the proposed network, we generated single-look ($L=1$) synthetic speckle images using optical images following equations \ref{eq1} and \ref{eq2}. We used the Berkeley segmentation dataset (BSD) \cite{BSD} to obtain the optical images where 450 and 50 images of size $256 \times 256$ were allocated for training and validation, respectively. The proposed network was implemented using PyTorch and trained with a learning rate of 0.0002 for 400 epochs. The performance of the proposed algorithm on synthetic speckled images were tested using a set of well-known testing images \cite{testdata} in terms of Peak Signal-to-Noise ratio (PSNR) and Structured Similarity Index (SSIM). We have compared the performance of the proposed network with PPB \cite{ppb}, SAR-BM3D \cite{sar_bm3d}, SAR-CNN \cite{SARCNN} and ID-CNN \cite{IDCNN}. Note that the two CNN-based methods (SAR-CNN and ID-CNN) were trained on the same synthetic data as the proposed method. 

The corresponding results are summarized in Table \ref{results_synthetic}. 
From Table \ref{results table}, it can be observed that the proposed method outperforms both traditional and CNN-based despeckling methods in terms of PSNR and SSIM when tested on synthetically speckled images.


\begin{table}[tb]
\setlength\tabcolsep{0pt}
\caption{Results on synthetic images of Set12 dataset \cite{testdata}.} \label{results table}
\centering
\smallskip
\begin{tabular*}{\columnwidth}{@{\extracolsep{\fill}}ccc}
\toprule
Method & PSNR & SSIM  \\
 \midrule
  PPB \cite{ppb}& 21.90 & 0.599  \\
  SAR-BM3D \cite{sar_bm3d}& 23.51 & 0.701 \\
  SAR-CNN \cite{SARCNN}& 24.51 & 0.651\ \\
  ID-CNN \cite{IDCNN}& 24.44 & 0.685\\
  Proposed method & \textbf{24.56} & \textbf{0.718}\\
\bottomrule
\label{results_synthetic}
\end{tabular*}
\end{table}

\begin{table}[h]
\setlength\tabcolsep{0pt}
\vspace{-1pt}
\caption{Results on real SAR images.} 
\label{results real table}
\centering
\smallskip
\begin{tabular*}{\columnwidth}{@{\extracolsep{\fill}}lSSSSSSSS}
    \toprule
    \multirow{2}{*}{Method} &
      \multicolumn{2}{c}{Region 1} &
      \multicolumn{2}{c}{Region 2} &
      \multicolumn{2}{c}{Region 3} &
      \multicolumn{2}{c}{Region 4}\\
      & {ENL $\uparrow$} & {Cx $\downarrow$} & {ENL$\uparrow$} & {Cx$\downarrow$} & {ENL$\uparrow$} & {Cx$\downarrow$} & {ENL$\uparrow$} & {Cx$\downarrow$}\\
      \midrule
    PPB  & \text{87.0} & \text{0.11}  & \text{125.5} & \text{0.09} & \text{21.0} & \text{0.22} & \text{117.8} & \text{0.09} \\
    SARBM3D&  \text{110.8} & \text{0.09} & \text{104.2} & \text{0.10} & \text{34.9} & \text{0.17} & \text{122.9}  &\text{0.09}\\
    SARCNN & \text{87.4} & \text{0.11} & \text{51.12} & \text{0.14} & \text{30.7} & \text{0.18} & \text{68.6}& \text{0.12}\\
    IDCNN & \text{47.6} & \text{0.14} & \text{33.3} & \text{0.17} & \text{23.1} & \text{0.21} & \text{34.5} & \text{0.17}\\
    Proposed & \textbf{154.4} & \textbf{0.08} & \textbf{171.6} & \textbf{0.08} & \textbf{39.2} & \textbf{0.16} & \textbf{133.19} & \textbf{0.08}\\
    
    \bottomrule
  \end{tabular*}
\end{table}

\begin{figure*}[h]
\centerline{\includegraphics[width=0.9\textwidth]{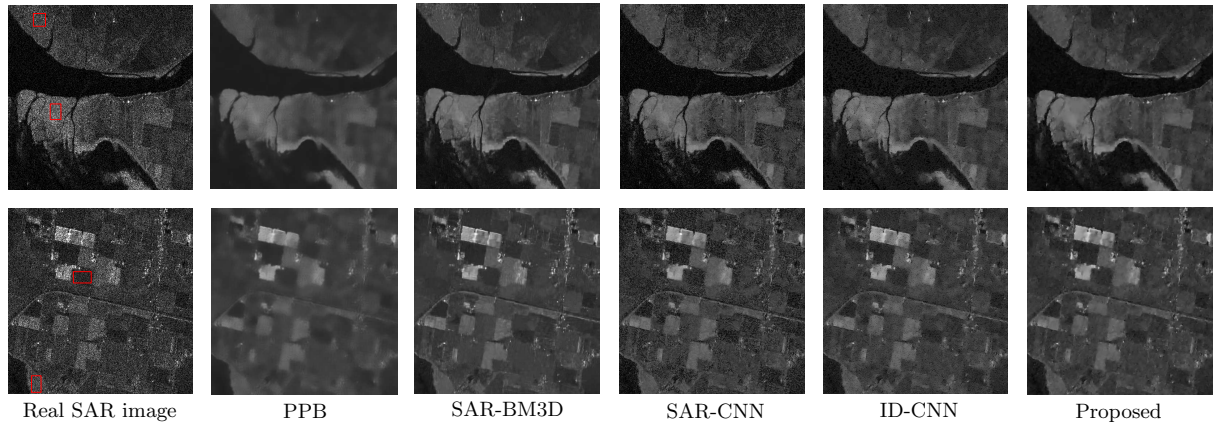}}
\vskip-10pt\caption{Results on real SAR images.}
\label{real_results_img}
\end{figure*}

We also evaluate the despeckling performance of our proposed method by testing on real SAR images. We compare the performance with the same despeckling approaches explained above. The despeckled results on the real SAR images are visualized in Fig. \ref{real_results_img} for qualitative comparison. Since real SAR images do not have clean ground truth images, we use equivalent number of looks (ENL) and the coefficient of variation (Cx) derived in a homogeneous region (regions considered in this study are marked as red boxes in Fig. \ref{real_results_img}) as quantitative measures for comparison. ENL is the ratio between the square of the mean and the variance of a homogeneous region, where as Cx is given by the ratio between standard deviation and the mean intensity of a homogeneous region. The results in terms of ENL and Cx are given in Table \ref{results real table}. Our proposed method resulted in the highest ENL values in all 4 regions which signifies the best despeckling performance out of the considered approaches. Low Cx values indicate better preservation of texture and our proposed algorithm gave the lowest Cx values in all cases. From Fig. \ref{real_results_img}, we can observe that ENL and Cx results are consistent with the visual results.


\section{Conclusion}
\label{sec:conclusion}
We proposed a network architecture that encompasses a transformer-based encoder and a convolution-based decoder, for SAR image despeckling. When compared with several existing filter-based and CNN-based despeckling methods, the results on synthetic and real SAR images show promising quantitative and qualitative improvements. The proposed method proved to be effective in reducing speckle  while preserving the texture and fine details in real SAR images.

\bibliographystyle{IEEEbib}
{\small{
\bibliography{refs}
}}
\end{document}